\documentclass[sigconf]{acmart}
\usepackage{graphicx}
\usepackage{subcaption}
\usepackage{cleveref}
\usepackage{multirow}
\AtBeginDocument{%
  }

\setcopyright{acmcopyright}
\copyrightyear{2018}
\acmYear{2018}
\acmDOI{XXXXXXX.XXXXXXX}

\acmConference[Conference acronym 'XX]{Make sure to enter the correct
  conference title from your rights confirmation emai}{June 03--05,
  2018}{Woodstock, NY}
\acmPrice{15.00}
\acmISBN{978-1-4503-XXXX-X/18/06}

\begin{document}

\title{Multimodal Fake News Detection via CLIP-Guided Learning}




\author{Yangming Zhou}
\affiliation{%
  \institution{School of Computer Science}
  \streetaddress{Fudan University}
  \city{Shanghai}
  \country{China}}
\email{ymzhou21@fudan.edu.cn}

\author{Qichao Ying}
\affiliation{%
  \institution{School of Computer Science}
  \streetaddress{Fudan University}
  \city{Shanghai}
  \country{China}}
\email{qcying20@fudan.edu.cn}

\author{Zhenxing Qian}
\authornote{Corresponding author. This work is supported by National Natural Science Foundation of China under Grant U20B2051, U1936214.}
\affiliation{%
  \institution{School of Computer Science}
  \streetaddress{Fudan University}
  \city{Shanghai}
  \country{China}}
\email{zxqian@fudan.edu.cn}

\author{Sheng Li}
\affiliation{%
  \institution{School of Computer Science}
  \streetaddress{Fudan University}
  \city{Shanghai}
  \country{China}}
\email{lisheng@fudan.edu.cn}

\author{Xinpeng Zhang}
\affiliation{%
  \institution{School of Computer Science}
  \streetaddress{Fudan University}
  \city{Shanghai}
  \country{China}}
\email{zhangxinpeng@fudan.edu.cn}

\renewcommand{\shortauthors}{Anonymous et al.}

\begin{abstract}
Multimodal fake news detection has attracted many research interests in social forensics. Many existing approaches introduce tailored attention mechanisms to guide the fusion of unimodal features. However, how the similarity of these features is calculated and how it will affect the decision-making process in FND are still open questions. Besides, the potential of pretrained multi-modal feature learning models in fake news detection has not been well exploited. This paper proposes a FND-CLIP framework, i.e., a multimodal Fake News Detection network based on Contrastive Language-Image Pretraining (CLIP). Given a targeted multimodal news, we extract the deep representations from the image and text using a ResNet-based encoder, a BERT-based encoder and two pair-wise CLIP encoders. The multimodal feature is a concatenation of the CLIP-generated features weighted by the standardized cross-modal similarity of the two modalities. The extracted features are further processed for redundancy reduction before feeding them into the final classifier. We introduce a modality-wise attention module to adaptively reweight and aggregate the features. We have conducted extensive experiments on typical fake news datasets. The results indicate that the proposed framework has a better capability in mining crucial features for fake news detection. The proposed FND-CLIP can achieve better performances than previous works, i.e., \textbf{0.7\%, 6.8\%} and \textbf{1.3\%}improvements in overall accuracy on Weibo, Politifact and Gossipcop, respectively. Besides, we justify that CLIP-based learning can allow better flexibility on multimodal feature selection.

\end{abstract}



\maketitle
\section{Introduction}
Online social networks have largely replaced the conventional way of information communication represented by newspapers and magazines. People enjoy the convenience of online social media in seeking friends or sharing viewpoints. 
However, OSNs have also promoted the wide and rapid spreading of fake news~\cite{FND-Survey,ICASSP-1,MVAE}. Online news posts can be more easily manipulated compared to written materials. News forgery can take various forms, for example, replacing a critical object within a picture with another one, or making biased or even misleading comments on the picture. What's worse, the readers are susceptible to well-crafted fake news and further circulate them. In sum, fake news is likely to create panic and misdirect public opinion, which alters society in negative ways. 

In the past decades, Fake News Detection (FND) has been the center of data-centric research for decades~\cite{allein2021like,shu2020fakenewsnet,xue2021detecting}. While manual observation towards all news and posts on the Internet is both expensive and time-consuming, automatic FND using machine learning is an efficient way to combat the widespread dissemination of fake news. FND has helped news readers identify bias and misinformation in news articles and therefore stop their spreading. 
Early works on fake news detection merely focused on text-only or image-only content analysis~\cite{WWW-4,Leveraging_6}. A pretrained model is usually employed to verify the logical and semantic soundness of the input. Also, trivial clues such as grammatical errors or traces left by image manipulation might be taken into consideration.
While unimodal FND schemes are effective, modern news and posts are usually with rich information of several modalities and these methods neglect their correlation. 
For some fake news, a real image can be combined with total rumors and correct words can be used to describe a tampered image. 
In that sense, multimodal feature analysis is required to offer complementary benefits to assist FND. 

\begin{figure}[!t]
  \centering
  \includegraphics[width=0.95\linewidth]{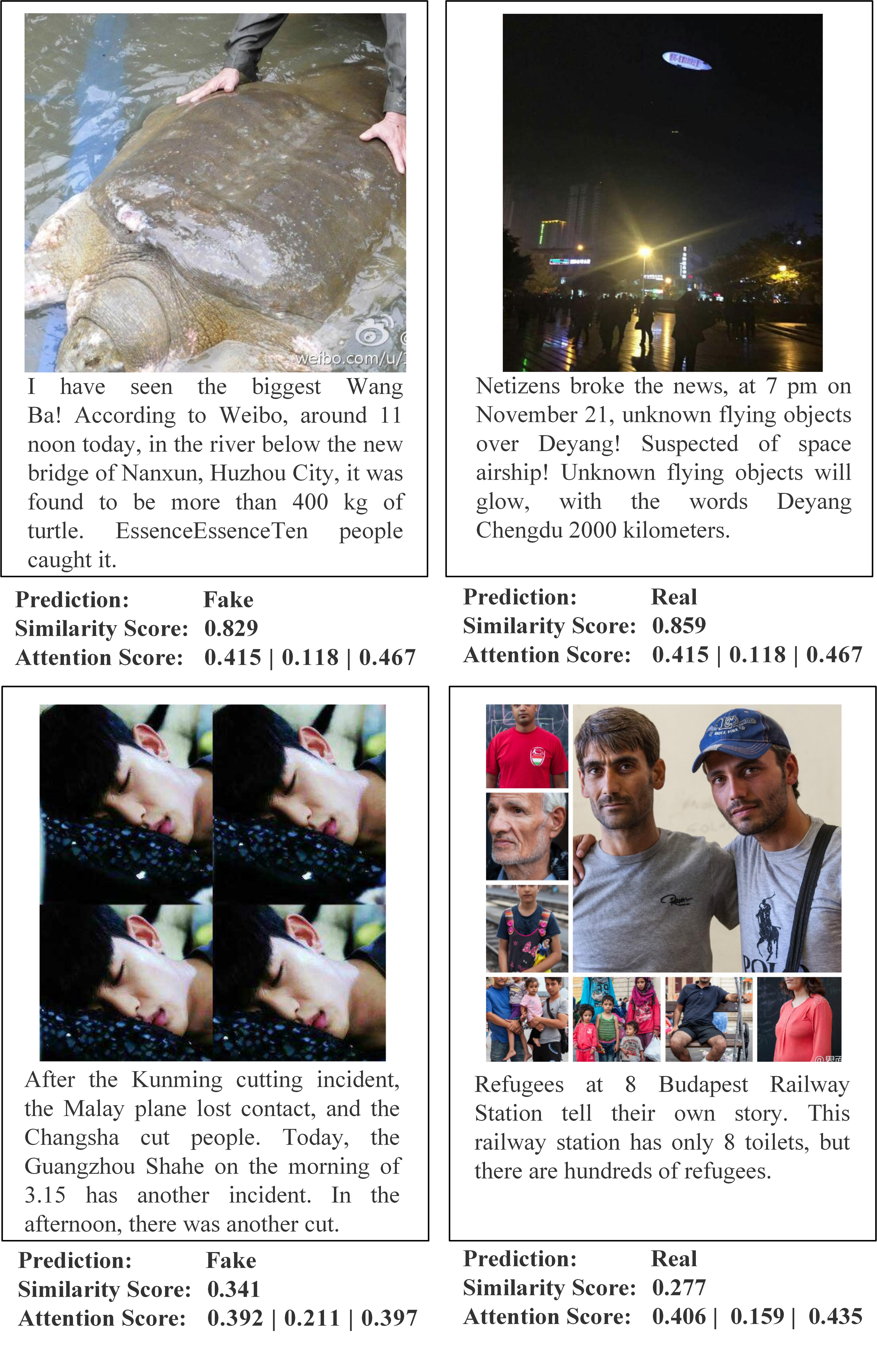}
  \caption{Examples of fake news detection using FND-CLIP.The three attention scores of each news are text score, image score, and fusion score respectively.}
  \label{intro}
\end{figure}

In recent years, there have already been a lot of works that aggregate multimodal features to detect anomalies in news and posts~\cite{MVAE,EANN,WWW}.
Besides fusing features from images and texts, comments, up-vote ratio and the spreading graph are mostly preferred by researchers to evaluate the truthfulness of a post. These additional modalities are interactive and change over time. 
Many previous works prefer using as much modalities as possible.
However, interactive modalities are less dependable than images and texts, or static modalities.
First, the absence of interactive modalities might be common. A typical example is that no clue can be left in historical behavior in news posted by newly registered users, and nor can it be left in comments or up-votes if we wish to reject fake news shortly after their submission. 
Second, interactive modalities are less stable and can be changed over time, therefore potentially resulting in varied forensics results. 
Therefore, we revisit current arts in FND with only static modalities, and find that though many algorithms design well-crafted networks for multimodal feature fusion~\cite{SpotFake,EANN}, the mechanisms are largely at a black-box level as to how multimodal features will influence the final decision. 
Some works try to address this issue by explicitly calculating correlation on generating fused features. For example, Chen et al.~\cite{WWW} additionally train variational auto encoders (VAE) that first compress the images and texts and contrastively learns to minimize the Kullback-Leibler (KL) divergence for news with correct image-text pairs. The corresponding cross-modal ambiguity score is then used to reweight the multimodal features~\cite{WWW}.  
Dhruv et al.~\cite{MVAE} propose MVAE that trains a decoder to reconstruct the original texts and low-level image features from the fused features. 
These methods have achieved decent performance in multimodal fake news detection.

However, there are still some issues for multimodal FND to be addressed. First, we find that how the similarity of features from different modalities is to be calculated and how it will affect the decision-making process in FND is still an open question. For \cite{WWW}, we are not sure how efficient the VAEs are so that the KL divergence will be small given matched image-text pairs. For MVAE, though the ability of reconstruction means that the fused features are able to contain more information, the necessity of these auxiliary tasks in the view of FND remains unknown.
Besides, we find that more advanced multimodal learning paradigms and pretrained models are not properly applied in FND. For example, CLIP~\cite{CLIP} is a multimodal model that combines knowledge of language concepts with semantic knowledge of images. It was trained on a variety of image-text pairs to predict the most relevant text snippet, given an image, and vice versa. CLIP, together with other advanced multimodal technologies can be beneficial in image-text feature fusing, yet their usages in FND still remain ill-posed.

This paper proposes FND-CLIP, a multimodal fake news detection network based on the pretrained Contrastive Language-Image Pretraining (CLIP) model. The CLIP-based learning for fake news detection is to address the issue of cross-modal ambiguity by explicitly measuring the correlation between texts and images of targeted posts, and to guide the feature fusing and decision-making stages. 
Specifically, we encode the image using a fine-tunable ResNet~\cite{RESNET} encoder a pretrained CLIP image encoder. The text is encoded by a fine-tunable BERT~\cite{BERT} encoder as well as a CLIP text encoder. The unimodal features are generated by concatenating the CLIP-generated features with the fine-tunable counterparts. The fused features consist of the two CLIP outputs. We use three projection heads to individually process the unimodal and fused features, which shrinks their sizes in order to distill the most important features for FND. Besides, we calculate the cosine similarity on the CLIP outputs and standardize it as the cross-modal similarity score. The score reweights the fused feature, where we regulate that less information will be provided by the fused features if the image and text show low correlation. Furthermore, we introduce an attention layer that outputs three scores that adaptively measure the significance of these features in their contribution to fake news detection. The classifier finally processes the summarized features to distinguish fake news from real ones. 

We have conducted extensive experiments on FND-CLIP on several typical datasets for FND, including a Chinese dataset named Weibo, and two English datasets named Politifact and Gossip. The results show that FND-CLIP achieves \textbf{0.7\%, 6.8\%} and \textbf{1.3\%} performance improvement in overall accuracy on the three datasets. Besides, we justify that CLIP-based learning can allow better flexibility on multimodal feature selection. 
Figure~\ref{intro} showcases four examples of fake news detection using FND-CLIP, where we see that the attention score as well as cross-modal similarity vary among different news instances. FND-CLIP is able to pay less attention to the multimodal features when the similarity is low, therefore flexibly aggregating information according to the characteristics of the provided news.

The contributions of this paper are mainly three-folded, namely:
\begin{itemize}
\item We propose FND-CLIP, a multimodal fake news detection method with CLIP-based learning, where the CLIP pretrained model is used to measure the cross-modal similarity and guide the mapping and fusion of features.
\item We propose a modality-wise attention mechanism to adaptively weight the text, image, and fused features. Given different news instances, we find that the model flexibly learns to pay more attention to useful information in unimodal or multimodal features.
\item We have conducted comprehensive experiments on three famous datasets, where the results prove that CLIP-generated features can be important assists to the unimodal features. FND-CLIP outperforms state-of-the-art fake news detection methods. 
\end{itemize}


\section{Related Works}
\subsection{Unimodal Fake News Detection}
Unimodal FND usually works on finding anomalies in either the text or the image of a post.  These algorithms often follow the essence of human decision process. For images, Cao et al.~\cite{Leveraging_5} jointly study image forensics features, semantic features, statistical features, and context features for fake news detection. It suggests that typical methods for image manipulation detecion~\cite{MVSS} are useful in unveiling traces for news tampering. Besides, semantic inconsistency regarding the common sense~\cite{li2021entity} as well as poor image quality~\cite{han2021fighting} can be widely present in fake news. For texts, verifying the logical soundness is essential~\cite{guo2018rumor}, also accompanied by finding clues such as grammatical errors, writing styles~\cite{Leveraging_27} or extracting rhetorical structure~\cite{Leveraging_6}. 
Besides, both linguistic and visual patterns can be highly dependent on specific
events and corresponding domain knowledge. Therefore, Nan et al.~\cite{MDFEND} propose to employ domain gate to aggregate multiple representations extracted by mixture-of-experts, and it deals with multi-domain fake news propagation in the language modality.

Though these unimodal characteristics can be explored and they indeed play key roles in distinguishing fake news, the multimodal characteristics such as correlation and consistency are ignored, which potentially impair the overall performance of these unimodal schemes on multimodal news. 

\begin{figure*}[!t]
  \centering
  \includegraphics[width=1.0\linewidth]{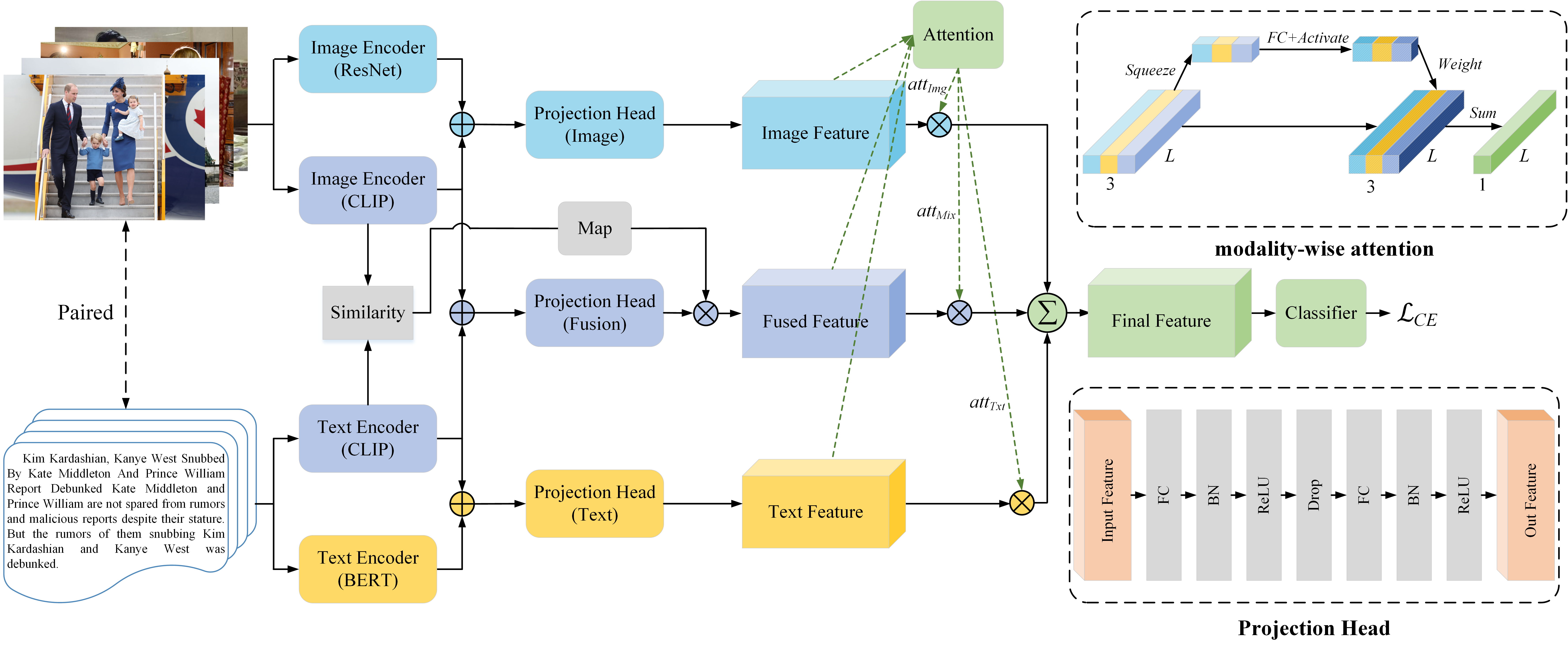}
  
  \caption{The architecture of the proposed FND-CLIP method. CLIP, BERT, and ResNet are used to extract the features of different modalities of multimodal news. Encoded features of different levels are obtained through projection heads. CLIP similarity score is calculated to determine the importance of fused feature. A modality-wise attention mechanism is further used to reweight different modal features adaptively for the classifier to classify fake news.}
  \label{architecture}
\end{figure*}

\subsection{Multimodal Fake News Detection}
In the past literature, many works have been done on mining useful representations from images and texts of the news for fake news detection. 
Earlier works design sophisticated yet black-box attention mechanisms for multimodal feature fusion~\cite{WWW-4,WWW-3}.
Many other works~\cite{EANN,MVAE,WWW} propose to better align the extracted features from different modalities before sending them into the classifier.
Wang et al.~\cite{EANN} propose EANN that further employs an auxiliary task of event classification to aid feature extraction. The event classification branch is designed to better disentangle the mined multimodal features so that there are both event-specific information and event-agnostic information.
Dhruv et al.~\cite{MVAE} processe the image and text using unimodal feature extractors and further utilize a multimodal VAE to learn a shared representation from them.  The sampled representation produced by the VAE is then sent to a decoder which tries to reconstruct the original texts and low-level image features. 
Besides the focus on network design, other works exploit more information from the datasets. For example, Qi et al.~\cite{Entity-Enhanced} claim that image feature extractors cannot well understand visual entities such as celebrities, landmarks, and texts within the images, and therefore propose to manually extract these kinds of information as linguistic assists. Zhang et al.~\cite{Dual-Emotion} design a novel dual emotion feature descriptor to measure the emotional gap between the post and its comments and verify that dual emotion is distinctive between fake and real news. Chen et al.~\cite{WWW} use two VAEs to compress the images and texts and contrastively learn to minimize the Kullback-Leibler (KL) divergence for correctly matched image-text pairs. The resultant score is then used to reweight the multimodal features during feature fusing. 

Though these methods have achieved decent performance in multimodal FND, there are still issues to be concerned. First, how to explicitly measure the correlation between images and texts within a post still remain unclear. Second, we see that little work in FND consider applying the recently emerged arts in multimodal learning, which motivates us to use the CLIP-based pretraining to further boost the performance.

\subsection{Multimodal Learning}
Recent years have shown rapid developments in the field of multimodal machine learning~\cite{Multimodal-Review}. 
Neural architectures are employed in tasks that go beyond single modalities, for example, Visual Question Answering (VQA)~\cite{VQA}, Visual Commonsense Reasoning (VCR)~\cite{VCR}, etc. 
In these tasks and beyond, priors and features from different modalities are required and algorithms or deep networks cannot be effective when provided with only a single modality.
Several generic technologies are developed for learning joint representations of image content and natural language. For example, the CLIP model~\cite{CLIP} is designed as a bridge between computer vision and natural language processing. It was trained on a variety of image-text pairs to predict the most relevant text snippet, given an image, without directly optimizing for the task. The model consists of two encoders that respectively embed texts and images into a uniform mathematical space. Then, for the matched image-text pair, CLIP is encouraged to maximize the cosine similarity between the embedding of the two modalities. Otherwise, the similarity is minimized for the model to find the most suitable paired images and texts. 
Multimodal learning has a promising future where the innovation of CLIP has benefitted a lot of down-stream tasks~\cite{Hair-CLIP,CLIP-Art}.
Other multimodal schemes can be represented by Glide~\cite{Glide} and VilBERT~\cite{vilbert} that are respectively for text-to-image generation and multimodal representation learning.


\section{METHOD}
\subsection {Approach Overview}
For multimodal fake news detection, we collect the statc modalities of the sampled news that includes text and image, and denote each sample as $\mathbf{x}=({\mathbf{x}_{Txt}},{\mathbf{x}_{Img}})$. The ground-truth label is $y$ where $y=0$ indicates that $\mathbf{x}$ is a real news, otherwise $y=1$. 
According to the most traditional multimodal learning paradigm, a rich set of features are first extracted from ${\mathbf{x}_{Txt}}$ and ${\mathbf{x}_{Img}}$ that both represents the unimodal characteristics and the multimodal characteristics, which are then further fused and projected into a single value of $\hat{y}$ that should be close to the ground truth.
\begin{equation}
 \hat{y}={{F}_{cls}}({{F}_{Mix}}({{F}_{Txt}}(\mathbf{x}_{Txt}),{{F}_{Img}}(\mathbf{x}_{Img}))),
\end{equation}
where ${{F}_{Txt}}$ and ${{F}_{Img}}$ are unimodal feature extractors, ${F}_{Mix}$ is the feature fusing model and ${F}_{cls}$ is the classification head. 
In order to model ${{F}_{Txt}}$ and ${{F}_{Img}}$, most of the previous methods use different pre-trained models to extract text and image features in different semantic spaces, and for ${{F}_{Mix}}$, the proposed mechanisms vary. The crucial point is how to ensure that features provided from both modalities will be utilized in the later stage, otherwise the gap in semantic space makes the fused features unable to accurately represent the correlation between image and text. 
Instead of applying sophisticated and black-box feature-fusing networks, we employ a simple yet effective method where pretrained networks for multimodal learning is introduced to extract aligned multimodal features and to guide the learning of the classification network. We choose the CLIP model~\cite{CLIP} to measure the cross-modal similarity considering that the model is trained to provide the most appropriate language description of a given image and vice versa, and therefore is in line with the above requirements. After feature extraction and alignment, we use a light-weight network to implement ${L}_{Cls}$ which predicts $\hat{y}$.\par

\subsection {Network Specification}
Figure~\ref{architecture} illustrates the network design of FND-CLIP.
The whole pipeline consists of four main modules, namely, unimodal feature encoder, CLIP-based encoder, projection and attention module, and finally the classifier. \par

\noindent\textbf {Unimodal feature generation.}
We use a pretrained BERT model to obtain the feature ${{f}_{\emph{BERT}}}\in {{\mathbb{R}}^{n_\emph{BERT}}}$  of ${\mathbf{x}_{Txt}}$. 
For the image ${{\mathbf{x}}_{Img}}$, we use ResNet~\cite{he2016deep} to get deep representations ${{f}_{\emph{ResNet}}}\in {{\mathbb{R}}^{n_\emph{ResNet}}}$ from the image.
Besides ${{f}_{\emph{BERT}}}$ and ${{f}_{\emph{ResNet}}}$, we use CLIP encoders to encode text and image and obtain the features ${{f}_{\emph{CLIP-T}}}\in {{\mathbb{R}}^{n_\emph{CLIP}}}$ and ${{f}_{\emph{CLIP-I}}}\in {{\mathbb{R}}^{n_\emph{CLIP}}}$.
In order to improve the representation capability of the unimodal branches, embedding concatenation are performed in the text and image unimodal intra-modalities, respectively,

\begin{equation}
\left\{\begin{array}{l}
{{f}_{Txt}}={\emph{concat}}\left({{f}_{\emph{BERT}}},{{f}_{\emph{CLIP-T}}}\right) \\
{{f}_{Img}}={\emph{concat}}\left({{f}_{\emph{ResNet}}},{{f}_{\emph{CLIP-I}}}\right) ,
\end{array}\right.
\end{equation}
where ${{f}_{Txt}}\in {{\mathbb{R}}^{n_\emph{BERT}+n_\emph{CLIP}}}$ and ${{f}_{Img}}\in {{\mathbb{R}}^{n_\emph{ResNet}+n_\emph{CLIP}}}$.


\noindent\textbf {CLIP-guide multimodal feature generation.} 
The text and image features extracted by BERT and ResNet respectively have significant cross-modal semantic gaps, and it is difficult for the network to learn their intrinsic semantic correlation if they are fused directly. Therefore, the two features are only used as unimodal representation, while the multimodal representation is obtained by first concatenating the alignment features of the text-image pair extracted by CLIP and then fine-tuning them to reduce redundancy and introduce attention. 
The concatenated feature is denoted as ${{f}_{Mix}}\in {{\mathbb{R}}^{2 \times n_\emph{CLIP}}}$, where
\begin{equation}
{{f}_{Mix}}=\emph{concat}({{f}_{CLIP-T}},{{f}_{CLIP-I}}).
\end{equation}
The multimodal features reflect the correlation between the two modalities and contain meaningful semantic information. The assistance of the multimodal features to unimodal features is to learn the cross-modal similarity. Previous works often use a single network to mine both coarse and fine features from a modality, which is quite demanding on the learning ability of the model. Here, with the introduction of CLIP model, BERT and ResNet, which is the pretraining models for unimodal tasks, can pay more attention to trivial clues compared to extracting semantic information. For example, BERT can better extract emotional features of texts, and ResNet can identify higher-frequent noise patterns of images. In contrast, the training strategy of CLIP uses large-scale image-text pairs to learn the extraction of semantics, while largely ignoring emotion, noise and other features irrelevant to image and text matching. Therefore, using CLIP for multimodal feature generation can well collaborate with the unimodal features to respectively scrutinize the news from different aspects.

After we get the three features of different modalities, we use three individual projection head ${P}_{\emph{Txt}}, {P}_{\emph{Img}}$ and ${P}_{\emph{Mix}}$ made up of Multi-Layer Perceptrons (MLP) to process the features. 
The goal is to reduce the dimension of the coarse features provided by the encoders and help filtering out redundant information. These networks share the same architecture but do not share weights. As is shown in Figure~\ref{architecture}, every the projection head contains two sets of full connected layer with Batch Normalization~\cite{BN} layer, a ReLU activation function, and a dropout layer. 

Merely combining the CLIP-based features as the multimodal features cannot necessarily provide enough reliable information. 
The reason is that the authenticity of news is not completely correlated with image-text correlation. Some news posts, no matter real or fake, lack cross-modal relation or even semantic information. In that case, some instances require more emotion, noise, and other features, and the corresponding multimodal features might be noisy when the similarity is low and fully utilizing such information might impair the performance.
To address the ambiguity issue between multimodal features, we measure the cosine similarity between the text features and the image features provided by CLIP, to adjust the intensity of fused features. The cosine similarity is calculated as follows.

\begin{equation}
\emph{sim}=\frac{{{f}_{Txt}}\cdot {{({{f}_{Img}})}^{T}}}{\left\| {{f}_{Txt}} \right\|\left\| {{f}_{Img}} \right\|}.
\end{equation}

Then, we apply standardization and a Sigmoid functions to map the similarity into the
range  $[0-1]$.  The normalization is done by calculating the running status of mean and standard deviation during training, and subtract the running mean from $\emph{sim}$ and divide it with the running standard deviation. Compared to the contrastive learning paradigm, the normalization helps to calculate the similarity without comparing the news post with other instances. 

Thus, the process of obtaining the projected unimodal and multimodal features is as follows.

\begin{equation}
\left\{\begin{array}{l}
   {{m}_{Txt}}=P_{Txt}\left({{f}_{Txt}} \right)  \\
   {{m}_{Img}}=P_{Img}\left({{f}_{Img}} \right)  \\
   {{m}_{Mix}}=Sigmoid\left(Std\left( \emph{sim} \right) \right)\cdot P_{Mix}\left( {{f}_{Mix}} \right).  \\
\end{array} \right.
\end{equation}

\noindent\textbf {Feature aggregation using modality-wise attention.} 
We apply an attention mechanism to reweight the projected features before aggregating the features from different modalities using spatial addition.
Inspired by the Squeeze-and-Excitation Network (SE-Net)~\cite{hu2018squeeze}, we designed a modality-wise attention module as shown in Figure~\ref{architecture} to weight each feature adaptively. First, the three $L\times1 $  features are concatenated into one $L\times3 $  feature, where $L$ represents the length of the feature. Average pooling and maximum pooling are adopted to squeeze a $1\times3 $  vector via summation, corresponding to the initial weight of each channel. Then, the initial weight obtained in the previous step is sent into the two $3\times3 $  fully connected layers with GELU~\cite{GELU} activation function, and normalized into the range $[0-1]$ using Sigmoid functions respectively to obtain the attention weights $\emph{att}=\{\emph{att}_{Txt},\emph{att}_{Img},\emph{att}_{\emph{Mix}}\}$. Finally, the weights are multiplied respectively on ${m}_{t}, {m}_{i}$ and ${m}_{\emph{mix}}$, and a sum process is performed to obtain the $L\times1$ aggregated feature ${{m}_{\emph{Agg}}}$.
\begin{equation}
{m}_{\emph{Agg}}=\emph{att}_{Txt}\cdot\emph{m}_{Txt}+\emph{att}_{Img}\cdot\emph{m}_{Img}+\emph{att}_{Mix}\cdot\emph{m}_{Mix}.
\end{equation}
\noindent\textbf {Classification and objective function.} 
We feed the aggregated representation ${{m}_{\emph{Agg}}}$ into a two-layer fully-connected network as the classifier ${F}_{cls}$ to predict the label $\hat{y}$.
The objective function of FND-CLIP is to minimize the cross-entropy loss to correctly predict the real and fake news.
\begin{equation}
\mathcal{L}_{CE}=y \log \left( \hat{y}\right)+\left(1- y\right) \log \left(1- \hat{y}\right) \text {. }
\end{equation}

\begin{table*}[!t]
  \caption{\noindent\textbf{Performance comparison between FND-CLIP and other methods on three datasets.} Our method achieves the highest accuracy among these methods, and its precision, recall, and FI-score are also higher than most of the compared methods.}
  \label{comparison}
  \setlength{\tabcolsep}{3mm}{
\begin{tabular}{ccccccccc}
\toprule
\multirow{2}{*}   & \multirow{2}{*}{Method} &\multirow{2}{*}{Accuracy} &\multicolumn{3}{c}{Fake News}  & \multicolumn{3}{c}{Real News}\\
   \cline{4-9}&       &     & Precision   & Recall  & F1-score    & Precision    & Recall      & F1-score             \\
\midrule
\multirow{9}{*}{Weibo} 

    & EANN~\cite{EANN}   & 0.827 & 0.847 & 0.812     & 0.829     & 0.807       & 0.843      & 0.825        \\
& MVAE~\cite{MVAE}    & 0.824    & 0.854   & 0.769    & 0.809    & 0.802  & 0.875   & 0.837  \\
& Spotfake~\cite{singhal2019spotfake}    & 0.892    & 0.902  & \textbf{0.964}   & \textbf{0.932}   & 0.847    & 0.656    & 0.739   \\
 & MVNN~\cite{xue2021detecting}     & 0.846       & 0.809   & 0.857 & 0.832     & 0.879      & 0.837      & 0.858    \\
&SAFE~\cite{zhou2020mathsf}     &0.762   &0.831  &0.724  &0.774  &0.695  &0.811  &0.748 \\
&LIIMR~\cite{singhal2022leveraging}    & 0.900    & 0.882   & 0.823   & 0.847    & 0.908    & \textbf{0.941}   & \textbf{0.925}  \\
 & MCAN~\cite{wu2021multimodal}      & 0.899    & 0.913   & 0.889   & 0.901   & 0.884   & 0.909   & 0.897   \\
& CAFE~\cite{WWW}    & 0.840     & 0.855  & 0.830  & 0.842  & 0.825   & 0.851  & 0.837    \\
& FND-CLIP   & \textbf{0.907}  & \textbf{0.914} & 0.901  & 0.908& \textbf{0.914} & 0.901  & 0.907 \\
\midrule
\multirow{7.5}{*}{Politifact} 
& RoBERTa-MWSS~\cite{shu2020leveraging}   & 0.820    &  & -     & -  & 0.820   & -     & -    \\
  & SAFE~\cite{zhou2020mathsf}   & 0.874   & 0.851    & 0.830    &0.840   & 0.889   & 0.903   & 0.896   \\
 & Spotfake+~\cite{SpotFake}  & 0.846  & -   & -    & -   & -  & -   & -    \\
 & TM~\cite{TM}   & 0.871  & -    & -     & -   & 0.901 & -    & -   \\
& LSTM-ATT~\cite{lin2019detecting}   & 0.832  & 0.828  &0.832   & 0.830  & 0.836  & 0.832 & 0.829  \\
& DistilBert~\cite{allein2021like}  & 0.741    & 0.875 & 0.636  & 0.737 & 0.647 & 0.880  & 0.746 \\
& CAFE~\cite{WWW}  & 0.864 & 0.724 & 0.778 & 0.750 & 0.895 & 0.919    & 0.907  \\
& FND-CLIP & \textbf{0.942}   & \textbf{0.897} & \textbf{0.897}   & \textbf{0.897}    & \textbf{0.960} & \textbf{0.960}& \textbf{0.960}  \\
\midrule
\multirow{7.5}{*}{Gossipcop} 
& RoBERTa-MWSS~\cite{shu2020leveraging} & 0.800  & -    & -   & 0.800   & - & -   & - \\
& SAFE~\cite{zhou2020mathsf}  & 0.838  & 0.758  & 0.558   & 0.643     & 0.857  & 0.937  & 0.895  \\
 & Spotfake+~\cite{SpotFake}  & 0.856   & - & - & -    & -    & -    & -  \\
& TM~\cite{TM}  & 0.842 & - & -   & - & 0.896   & -  & -  \\
& LSTM-ATT~\cite{lin2019detecting}  & 0.842   & \textbf{0.845}   & \textbf{0.842}   & \textbf{0.844}  & 0.839  & 0.842 & 0.821  \\
& DistilBert~\cite{allein2021like}  & 0.857  & 0.805  & 0.527  & 0.637  & 0.866& \textbf{0.960}& 0.911\\
& CAFE~\cite{WWW}  & 0.867 & 0.732  & 0.490  & 0.587   & 0.887& 0.957  & 0.921 \\
& FND-CLIP & \textbf{0.880}  & 0.761 & 0.549   & 0.638       & \textbf{0.899}    & 0.959 & \textbf{0.928}    \\      
    \bottomrule
  \end{tabular}}
\end{table*}

\subsection{Training Detail} 
On the selection of BERT pretrained models, we respectively use the “bert-base-chinese” model on Chinese data and the “bert-base-uncased” model on English data, perform an attention-based post-processing~\cite{jawahar2019does}. The length of the input text is set to 300 words. About the ResNet, we use pre-trained ResNet-101 to extract  visual features, setting the size of the input image to 224 × 224.The size of the images inputted to CLIP is the same as that to ResNet. Science CLIP has not pre-trained Chinese text model, we use Google Translation API~\cite{johnson2012google} to translate Chinese texts to English. In addition, we use the summary generation model~\cite{raffel2019exploring} to generate summary statements as the CLIP input for the text with the size longer than 50, to meet the requirements that the input size of the text has an upper bound in CLIP. The used pre-trained CLIP model is “ViT-B/32”.  We fine-tune ResNet in training stage, while freezing the weights of BERT and CLIP due to their difficulty in training on small datasets. We implement the projection heads using two fully connected layers with 256 and 64 hidden units, respectively. The hidden sizes of the two fully connected layers in the classifier are 64 and 2, respectively. The batch size is set as 64.

We use Adam optimizer~\cite{kingma2014adam} with the default parameters. The learning rate is  $1\times {{10}^{-3}}$ where weight decay is 12. We trained a model for 50 epochs and chose the epoch getting the best test accuracy among them as the final result to avoid over-fitting.\par

 \section{EXPERIMENTS}

\subsection{Experimental Setup}
\noindent\textbf {Dataset.} We use three real-world datasets collected from social media, namely, Weibo~\cite{jin2017multimodal}, Gossipcop, and Politifact~\cite{shu2020fakenewsnet}. During experiments, the unimodal news posts with no image or no text description were filtered out. If a news post contains a text with multiple associated images, we randomly select one image.
Weibo is a widely used Chinese dataset in fake news detection. 
The training set contains 3, 749 real news and 3, 783 fake news, and the test set contains 1, 996 news. 
Politifact and Gossipcop datasets are two English datasets collected from the political and entertainment domains of FakeNewsNet~\cite{shu2020fakenewsnet} repository, respectively. 
Politifact contains 244 real news and 135 fake news in the training set and 75 real news and 29 news in the test set. 
Gossipcop contains 10, 010 training news, including 7, 974 real news and 2, 036 fake news. The test set contains 2, 285 real news and 545 fake news.
Besides, while Twitter~\cite{Twitter} is also a well-known multimodal dataset for FND, we find that it contains plenty of duplicated posts and over 10k posts host only 463 images. More importantly, more than 70\% of tweets on Twitter dataset are related to a single event, which can easily lead to model overfitting. Therefore, we do not conduct experiments on Twitter.

\begin{figure}[!t]
  \centering
  \includegraphics[width=1.0\linewidth]{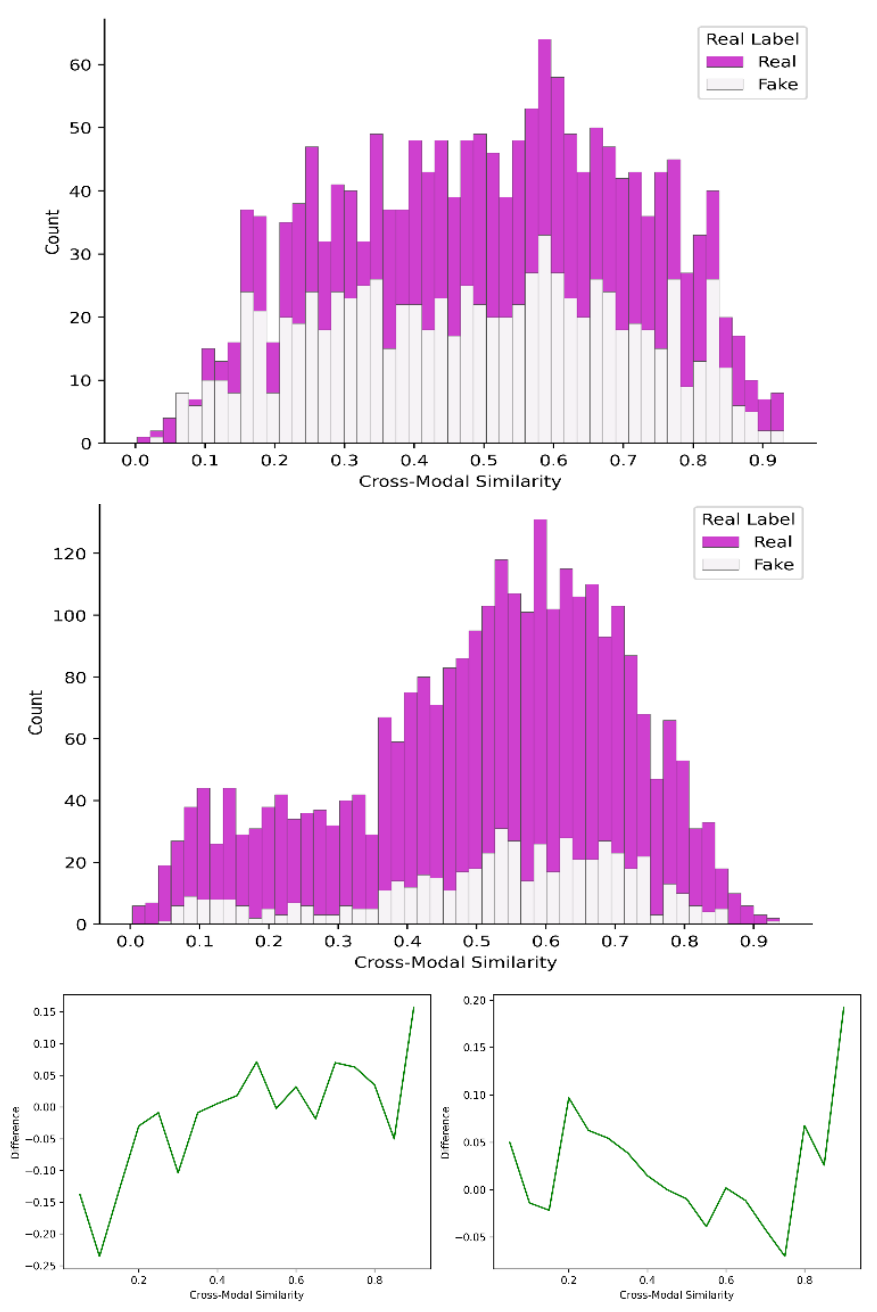}
  
  \caption{Statistical analysis on cross-modal similarity of different news. The first and second row respectively show the counting of real/fake news according to cross-modal similarity on Weibo and Gossipcop. The third row shows the distance between the real news rate in each bin compared to the average rate on the corresponding dataset (left: Weibo. right: Gossipcop).}
  \label{ambiguity}
\end{figure}
\noindent\textbf {Baseline Methods.}
For a fair and reproducible comparison, we have to be selective in choosing the baseline methods. First, we prefer methods that provide pre-trained models or source code publicly available. Second, the methods should follow a common evaluation
protocol where the three datasets are used for training and testing. Accordingly, we compare FND-CLIP with the following methods and provide a quick recap.

\textbf{EANN~\cite{EANN}}, which employs an auxiliary task of event classification to improve generalizability.

\textbf{MVAE~\cite{MVAE}}, which uses a variational autoencoder to model representations between text and images for fake news detection.

\textbf{Spotfake~\cite{singhal2019spotfake}}, which uses VGG and BERT to respectively extract image and text features and concatenates them to classify.

\textbf{MVNN~\cite{xue2021detecting}}, which incorporates textual semantic features, visual tampering features, and similarity of textual and visual information in fake news detection.

\textbf{SAFE~\cite{zhou2020mathsf}}, which fed the relevance between news textual and visual information into a classifier to detect fake news.

\textbf{LIIMR~\cite{singhal2022leveraging}}, which identifies and suppresses information from weaker modalities and extracts relevant information from the strong modality on a per-sample basis.

\textbf{MCAN~\cite{wu2021multimodal}}, which stacks multiple co-attention layers to fuse the multimodal features.

\textbf{CAFE~\cite{WWW}}, which formulates an ambiguity-aware multimodal fake news detection method to adaptively aggregate unimodal features and cross-modal correlations.

\textbf{RoBERTa-MWSS~\cite{shu2020leveraging}}, which exploits multiple weak signals from different sources from user and content engagements.

\textbf{Spotfake+~\cite{SpotFake}}, which is an improved version of Spotfake and can detect full length articles.

\textbf{TM~\cite{TM}}, which utilizes lexical and semantic properties of both true and fake news text to detect fake news.

\textbf{LSTM-ATT~\cite{lin2019detecting}}, which builds a model based on XGBoost to detect full length fake news.

\textbf{DistilBert~\cite{allein2021like}}, which uses latent representations of news articles and user-generated content to guide model learning.

\subsection{Performance Analysis}
Table~\ref{comparison} shows the average precision, recall, and accuracy of FND-CLIP on three representative datasets. The results are promising, with over 90\% average accuracy on Weibo and over 94\% on Politifact, which indicates that the proposed method is a dependable and robust fake news detection algorithm that can detect anomalies given multi-lingual and multi-domain news. Especially, the recall rates of real news on the three datasets are all above 0.9, and therefore FND-CLIP is less likely to classify a real news as fake.

To further conduct statistical analysis on how cross-modal similarity correlates with the attention score and how they vary given different news instances, Figure~\ref{ambiguity} shows the correlation between the CLIP-based cross-modal similarity score and the fake news ratio on Weibo and Gossipcop dataset. In row 1 and 2, we group all news in each dataset into several bins according to their similarity score, and find that a news is more likely to be real when the similarity score is high. In row 3, we calculate real news rates of each bin and subtract them with the average real news rate of the corresponding dataset. The curves show that real news rate goes up with the increasing ambiguity on Weibo, and first goes down then surges up on Gossipcop. Such statistical charateristics are useful for deep networks to identify fake news.

\subsection{Comparison with State-of-the-arts}
We further compare FND-CLIP with the above-mentioned state-of-the-art methods and the comparison results are presented in Table~\ref{comparison}.
'-' means the results are not available from the original paper. As shown in Table~\ref{comparison}, FND-CLIP outperforms all the compared methods on the three datasets in terms of Accuracy, and achieves slightly lower than Spot on Weibo in Recall. FND-CLIP achieves the highest accuracy of 90.7\%, 94.2\%, and 88.0\%, which surpasses 0.7\%, 6.8\%, and 1.3\% over the state-of-the-art method, on the three real-world datasets, respectively. Besides, we rank either $1_{st}$ or $2_{nd}$ in precision, recall, and accuracy in all tests, which proves the effectiveness of FND-CLIP. \par
Many fake news detection methods, such as EANN and Spotfake, rely only on the fused features obtained by direct use of concatenating or attention mechanisms. However, these fused features cannot provide sufficient discrimination ability to classify fake news, because the text and image features separately extracted are not in the same semantic space and the correlation information of the text and image is not well-paid attention to during the fusion process. Therefore, the experimental results of these methods are unsatisfactory.
CAFE uses cross-modal alignment to train encoders that can map texts and images into the same semantic space. By using the features fused from the alimented text and image features to classify, it achieves good experimental results, especially on the Politifact and Gossiopcop datasets. However, due to the limitation of the number of data sets and the rough label method for training labels, the encoding effect of the encoder is not optimal, and the semantic gap between text and image features is still significant. In addition, CAFE designs an ambiguity learning module to calculate a weight used for adaptively adjusting the calculation of different modalities. However, the weights for selecting unimodal or multimodal features are obtained by manual calculation, and cannot be further optimized by reverse gradient propagation, thus affecting the performance of the detection.\par
FND-CLIP outperforms most of the state-of-the-art methods, mainly due to the following reasons. First, the pre-trained CLIP encoders in FND-CLIP can generate semantically information-rich text and image features in the same semantic space, ensuring the fused feature correctly reflects the correlation between text and image, and providing complementary information for the unimodal features. The modality-wise attention mechanism adaptively determines the weights of text, image, and fused features, avoiding the influence of invalid features on the representation ability of final features, and further improving the classification accuracy.

\begin{table*}[!t]
    \caption{Ablation study on the architecture design and different features of FND-CLIP on three datasets.The entire FND-CLIP achieved the highest accuracy and F1-score, demonstrating that every module in the architecture of our method is effectiveness and every modality is effectively utilized.}
\label{modal}
  \setlength{\tabcolsep}{3mm}{
\begin{tabular}{ccccccccc}
\toprule
\multirow{2}{*}   & \multirow{2}{*}{Method} &\multirow{2}{*}{Accuracy} &\multicolumn{3}{c}{Fake News}  & \multicolumn{3}{c}{Real News}\\
  \cline{4-9}  &     &  & Precision    & Recall    & F1-score   & Precision    & Recall    & F1-score   \\

\midrule
\multirow {7}{*}{Weibo}  
& FND-CLIP multimodal-only  & 0.817  &0.899  &  0.718    & 0.798      & 0.761 &0.917   & 0.832   \\
& FND-CLIP image-only  & 0.796  &0.862 &  0.711     & 0.779   &0.750  & 0.884     & 0.811   \\
& FND-CLIP text-only  & 0.872    & 0.906 & 0.833   & 0.868   & 0.842  & 0.911    & 0.875   \\

& FND-CLIP w/o C& 0.874 &0.895 &  0.851 & 0.872& 0.855& 0.898& 0.876\\
    & FND-CLIP w/o F& 0.893   & 0.925& 0.857  & 0.890 &0.864&  0.929 & 0.895\\
    & FND-CLIP w/o A& 0.897 &\textbf{0.936} & 0.855   & 0.893 &0.863 & \textbf{0.940}  & 0.900\\

& FND-CLIP  & \textbf{0.907}   &0.914 &\textbf{0.901}   & \textbf{0.908}  &\textbf{0.901}& 0.914 & \textbf{0.907} \\
\midrule
\multirow {7}{*}{Politifact} 
& FND-CLIP multimodal-only  & 0.903   & 0.807  & 0.862   & 0.833   & 0.944   & 0.919   & 0.932   \\
& FND-CLIP image-only  & 0.748   &  0.600& 0.310    & 0.409   &  0.773 & 0.919    & 0.840   \\
& FND-CLIP text-only  & 0.903  & 0.913  &0.724   & 0.808    & 0.900 & \textbf{0.973}  & 0.935   \\

& FND-CLIP w/o C& 0.893   & 0.875    & 0.724 & 0.793 &0.899 & 0.960 & 0.928\\
    & FND-CLIP w/o F& 0.903     &0.880  &0.759 & 0.815 & 0.910 & 0.960 & 0.934\\
    & FND-CLIP w/o A& \textbf{0.942}     & \textbf{0.926}   &0.862 & 0.893 & 0.947 & \textbf{0.973}  & \textbf{0.960} \\

& FND-CLIP  & \textbf{0.942}&  0.897& \textbf{0.897}   & \textbf{0.897}  & \textbf{0.960} & 0.960  & \textbf{0.960} \\
\midrule
\multirow {7}{*}{Gossipcop} 
& FND-CLIP multimodal-only  & 0.862  & 0.708  & 0.484    & 0.575    & 0.886  &  0.952  & 0.918   \\
& FND-CLIP image-only  & 0.814 & \textbf{1.000}     &  0.033   & 0.064    & 0.813   & \textbf{1.000}    & 0.897   \\
& FND-CLIP text-only  & 0.871   &  0.741  & 0.508  & 0.603    & 0.891  &   0.958 & 0.923   \\

& FND-CLIP w/o C& 0.870   & 0.745    &0.494& 0.594&0.888 & 0.960 & 0.923\\
    & FND-CLIP w/o F& 0.874    &0.723   & 0.562& 0.632&0.901& 0.949 & 0.924\\
    & FND-CLIP w/o A& 0.873   & 0.715   & \textbf{0.567} & 0.633 &\textbf{0.902} &0.946 & 0.923\\
    
& FND-CLIP  & \textbf{0.880} & 0.761& 0.549   & \textbf{0.638} & 0.899 & 0.959   & \textbf{0.928} \\
\bottomrule
\end{tabular}}
\end{table*}

\subsection{Ablation Studies}
We explore the influence of the key components in FND-CLIP by evaluating the performance of the model with varied and partial setups. In each test, we remove different components and train the models from scratch.
The compared variants of FND-CLIP are implemented as follows.
\begin{itemize}
\item FND-CLIP w/o A. We remove the modality-wise attention module and direct aggregate the three features to obtain final feature; 
\item FND-CLIP w/o F. We remove the fusion module and use two unimodal features to classify news; 
\item FND-CLIP w/o C. We remove all CLIP-related modules and only use BERT and ResNet to extract text and image features.
\item FND-CLIP multimodal-only: We remove the unimodal feature extractor, BERT and ResNet, and only use CLIP fused feature as final feature; 
\item FND-CLIP image-only: We remove the all text-related features and only use image feature extracted by ResNet to classify; 
\item FND-CLIP text-only: We only use BERT-extracting feature to complete the detection task without any visual information.
\end{itemize}

\noindent\textbf{Effectiveness of Each Component. }
First, we analyze the impact of different components in FND-CLIP for fake news detection. 
From the results shown in Table~\ref{modal}, we have the following observations:  1) FND-CLIP outperforms FND-CLIP w/o C, proving that CLIP can effectively provide discernable features for fake news detection task and significantly improve the accuracy of classification.  Although only intra-modal features can be used for classification, the lack of interaction between modalities makes the final features lack the ability to represent the intrinsic relationship between images and texts.  2) FND-CLIP outperforms FND-CLIP w/o F, indicating that although the unimodal branches contain the CLIP-coded features, the fused feature reflecting the correlation of text and image provides effective multimodal information for classifier. Meanwhile, FND-CLIP w/o F outperforms FND-CLIP w/o C, indicating that the complement to unimodal features using CLIP-coded features is effective. 3) FND-CLIP outperforms FND-CLIP w/o A on Weibo and Gossipcop, indicating that modality-wise attention can help FND-CLIP adaptively weight useful modalities. FND-CLIP w/o A directly fuses the features of different modalities, which may cause the final feature be affected by invalid information from a modality.

\noindent\textbf{Contributions from Different Modalities. }
The second set of experiments is to evaluate the classification performance of different modalities in fake news detection.
From Table~\ref{modal}, we draw some analysis as follows: 1) FND-CLIP image-only performs worst, especially on Gossipcop dataset, where the F1 score of fake news was almost zero, meaning that all news was judged real and the model had no classification ability at all. This shows that in fake news detection, simple visual information provides fewer classification clues than other modalities. 2) FND-CLIP multimodal-only achieves accuracy of 81.7\%, 90.3\%, and 86.2\% on Weibo, Politifact, and Gossipcop datasets respectively, but performs worse than FND-CLIP text-only on Weibo and Gossipcop datasets, indicating that the correlation information of images and texts can be used to classify fake news. However, the classification ability of fused feature is limited because news itself has modal irrelevance and ambiguity. In addition, CLIP-based fused features focus on the semantics of the text, while the BERT-based text features also extract emotional features that are helpful for fake news detection. 3) FND-CLIP text-only achieves the second-best results, indicating that only using text feature can basically complete the classification task for fake news. However, FND-CLIP outperforms FND-CLIP text-only, proving the visual feature can supplement classification information and the correct use of multimodal features is superior to using only unimodal features in fake news detection.

\begin{figure*}[htbp]
  \centering
    \begin{subfigure}{0.26\linewidth}
      \centering
        \includegraphics[height=1\linewidth]{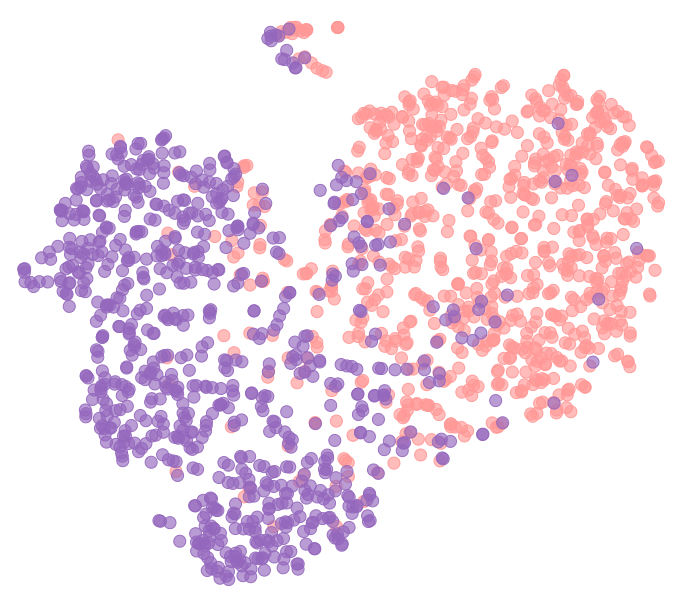}
        \caption{FND-CLIP}
        \label{FND-CLIP}
    \end{subfigure}
    \centering
    \begin{subfigure}{0.26\textwidth}
      \centering
        \includegraphics[height=1\linewidth]{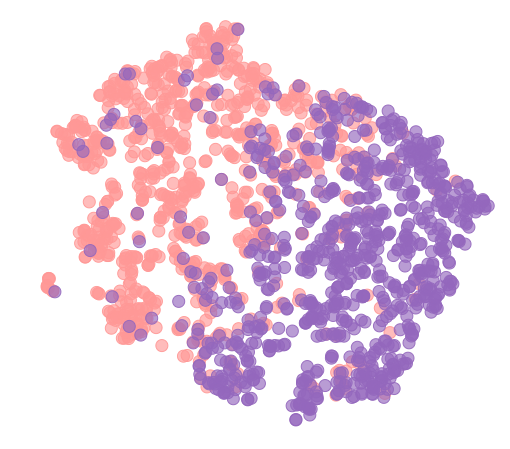}
        \caption{CAFE}
        \label{CAFE}
    \end{subfigure}
    \centering
    \begin{subfigure}{0.26\textwidth}
      \centering
        \includegraphics[height=1\linewidth]{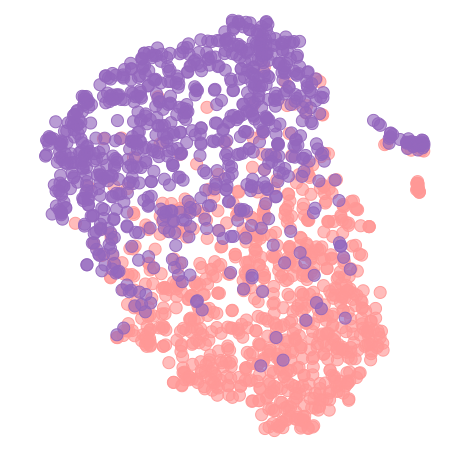}
        \caption{FND-CLIP w/o A}
        \label{FND-CLIP w/o A}
    \end{subfigure}
    \centering
    \begin{subfigure}{0.26\textwidth}
      \centering
        \includegraphics[height=1\linewidth]{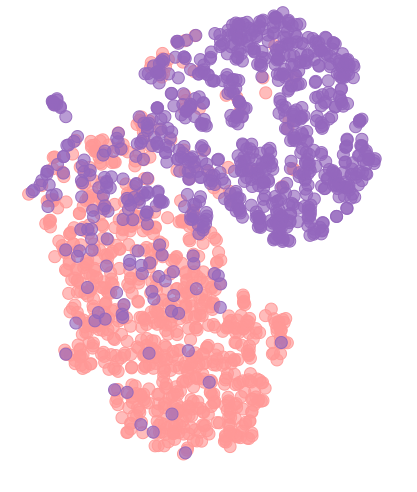}
        \caption{FND-CLIP w/o C}
        \label{FND-CLIP w/o C}
    \end{subfigure}
    \centering
    \begin{subfigure}{0.26\textwidth}
      \centering
        \includegraphics[height=1\linewidth]{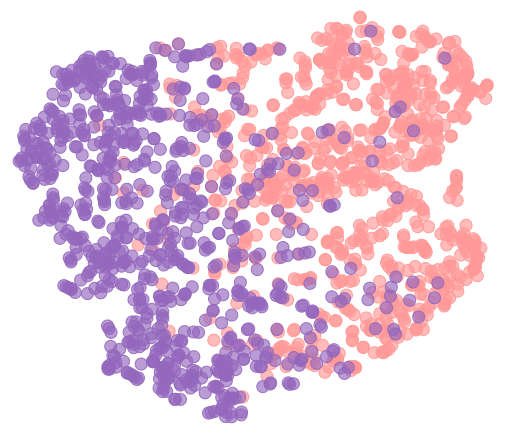}
        \caption{FND-CLIP text-only}
        \label{FND-CLIP text-only}
    \end{subfigure}
    \centering
    \begin{subfigure}{0.26\textwidth}
      \centering
        \includegraphics[height=1\linewidth]{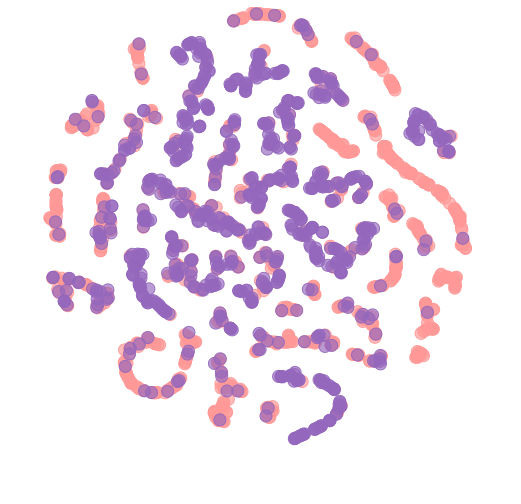}
        \caption{FND-CLIP image-only}
        \label{FND-CLIP image-only}
    \end{subfigure}
        \caption{T-SNE visualizations of the features before classifier that are learned by FND-CLIP , CAFE, FND-CLIP w/o C, FND-CLIP w/o A, FND-CLIP text-only, and FND-CLIP image-only on the test dataset of Weibo.}
        \label{T-SNE}
    \centering
    \label{tSNE}
\end{figure*}

\subsection{T-SNE Visualizations}
In Figure~\ref{tSNE}, we further analyze the proposed method using t-SNE~\cite{van2008visualizing} visualizations of the features before classifier that are learned by FND-CLIP, CAFE, and also the proposed method with partial settings such as FND-CLIP w/o C, FND-CLIP w/o A, FND-CLIP text-only, and FND-CLIP image-only on the test dataset of Weibo in Figure~\ref{T-SNE}.

The dots with the same color mean that they are within the same label. From Figure~\ref{T-SNE} we can see that the boundary of different label dots in FND-CLIP is more pronounced than that in CAFE,  FND-CLIP w/o C, and FND-CLIP w/o A, revealing that the extracted features in FND-CLIP are more discriminative than those in CAFE and the CLIP-related modules and modality-wise attention are useful for improving the classification ability of FND-CLIP. 

In addition, by comparing Figure~\ref{FND-CLIP}, Figure~\ref{FND-CLIP w/o C}, Figure~\ref{FND-CLIP text-only}, and Figure~\ref{FND-CLIP image-only}, we can see that image features alone are not enough for classification, which indicates that the image itself does not have classification ability. The effect of text-only is much better than that of image-only, Proving that the text features play a leading role in fake news detection, but there are still many sample dots that cannot be distinguished. FND-CLIP w/o C, which contains both text and image features, has a more obvious boundary of the dots than FND-CLIP text-only, indicating that different modalities have complementary information. In addition, the separation degree of the sample dots in Figure~\ref{FND-CLIP} is higher than that in Figure~\ref{FND-CLIP w/o C}, indicating that the multimodal features based on CLIP can improve the representation ability of the final features.



\section{Conclusions}

In this paper, we present a novel multimodal fake news detection method called FND-CLIP, which uses CLIP to extract aligned multimodal features and guide the learning of network for different modalities. In addition, we introduce modality-wise attention to adaptively determine the weights of text, image, and fused features. It can avoid introducing noisy and redundant features during feature fusion, which further improve the classification accuracy. 
We conduct comprehensive experiments on several well-known FND datasets.
The results show that using CLIP for multimodal feature generation can well collaborate with the unimodal features extracted by ResNet and BERT in mining crucial features for fake news detection.
More importantly, FND-CLIP outperforms many of the state-of-the-art methods in multimodal fake news detection.

Aside from the performance gain of FND-CLIP, the outputs are still in the form of binary value that predicts either ``real" or ``fake", which cannot somehow explain why the news is predicted fake and which elements in the news are most suspicious and abnormal.
In future works, we head towards developing more explainable fake news detection systems that can provide reasons why a given news is predicted as real or fake.


\bibliographystyle{ACM-Reference-Format}
\bibliography{CIKM}


\begin{thebibliography}{49}


\ifx \showCODEN    \undefined \def \showCODEN     #1{\unskip}     \fi
\ifx \showDOI      \undefined \def \showDOI       #1{#1}\fi
\ifx \showISBNx    \undefined \def \showISBNx     #1{\unskip}     \fi
\ifx \showISBNxiii \undefined \def \showISBNxiii  #1{\unskip}     \fi
\ifx \showISSN     \undefined \def \showISSN      #1{\unskip}     \fi
\ifx \showLCCN     \undefined \def \showLCCN      #1{\unskip}     \fi
\ifx \shownote     \undefined \def \shownote      #1{#1}          \fi
\ifx \showarticletitle \undefined \def \showarticletitle #1{#1}   \fi
\ifx \showURL      \undefined \def \showURL       {\relax}        \fi
\providecommand\bibfield[2]{#2}
\providecommand\bibinfo[2]{#2}
\providecommand\natexlab[1]{#1}
\providecommand\showeprint[2][]{arXiv:#2}

\bibitem[Allein et~al\mbox{.}(2021)]%
        {allein2021like}
\bibfield{author}{\bibinfo{person}{Liesbeth Allein},
  \bibinfo{person}{Marie-Francine Moens}, {and} \bibinfo{person}{Domenico
  Perrotta}.} \bibinfo{year}{2021}\natexlab{}.
\newblock \showarticletitle{Like Article, Like Audience: Enforcing Multimodal
  Correlations for Disinformation Detection}.
\newblock \bibinfo{journal}{\emph{arXiv preprint arXiv:2108.13892}}
  (\bibinfo{year}{2021}).
\newblock


\bibitem[Baltru{\v{s}}aitis et~al\mbox{.}(2018)]%
        {Multimodal-Review}
\bibfield{author}{\bibinfo{person}{Tadas Baltru{\v{s}}aitis},
  \bibinfo{person}{Chaitanya Ahuja}, {and} \bibinfo{person}{Louis-Philippe
  Morency}.} \bibinfo{year}{2018}\natexlab{}.
\newblock \showarticletitle{Multimodal machine learning: A survey and
  taxonomy}.
\newblock \bibinfo{journal}{\emph{IEEE transactions on pattern analysis and
  machine intelligence}} \bibinfo{volume}{41}, \bibinfo{number}{2}
  (\bibinfo{year}{2018}), \bibinfo{pages}{423--443}.
\newblock


\bibitem[Bhatt et~al\mbox{.}(2018)]%
        {WWW-3}
\bibfield{author}{\bibinfo{person}{Gaurav Bhatt}, \bibinfo{person}{Aman
  Sharma}, \bibinfo{person}{Shivam Sharma}, \bibinfo{person}{Ankush Nagpal},
  \bibinfo{person}{Balasubramanian Raman}, {and} \bibinfo{person}{Ankush
  Mittal}.} \bibinfo{year}{2018}\natexlab{}.
\newblock \showarticletitle{Combining neural, statistical and external features
  for fake news stance identification}. In \bibinfo{booktitle}{\emph{Companion
  Proceedings of the The Web Conference 2018}}. \bibinfo{pages}{1353--1357}.
\newblock


\bibitem[Bhattarai et~al\mbox{.}(2021)]%
        {TM}
\bibfield{author}{\bibinfo{person}{Bimal Bhattarai},
  \bibinfo{person}{Ole-Christoffer Granmo}, {and} \bibinfo{person}{Lei Jiao}.}
  \bibinfo{year}{2021}\natexlab{}.
\newblock \showarticletitle{Explainable Tsetlin Machine framework for fake news
  detection with credibility score assessment}.
\newblock \bibinfo{journal}{\emph{arXiv preprint arXiv:2105.09114}}
  (\bibinfo{year}{2021}).
\newblock


\bibitem[Bian et~al\mbox{.}(2020)]%
        {WWW-4}
\bibfield{author}{\bibinfo{person}{Tian Bian}, \bibinfo{person}{Xi Xiao},
  \bibinfo{person}{Tingyang Xu}, \bibinfo{person}{Peilin Zhao},
  \bibinfo{person}{Wenbing Huang}, \bibinfo{person}{Yu Rong}, {and}
  \bibinfo{person}{Junzhou Huang}.} \bibinfo{year}{2020}\natexlab{}.
\newblock \showarticletitle{Rumor detection on social media with bi-directional
  graph convolutional networks}. In \bibinfo{booktitle}{\emph{Proceedings of
  the AAAI conference on artificial intelligence}}, Vol.~\bibinfo{volume}{34}.
  \bibinfo{pages}{549--556}.
\newblock


\bibitem[Boididou et~al\mbox{.}(2018)]%
        {Twitter}
\bibfield{author}{\bibinfo{person}{Christina Boididou}, \bibinfo{person}{Symeon
  Papadopoulos}, \bibinfo{person}{Markos Zampoglou}, \bibinfo{person}{Lazaros
  Apostolidis}, \bibinfo{person}{Olga Papadopoulou}, {and}
  \bibinfo{person}{Yiannis Kompatsiaris}.} \bibinfo{year}{2018}\natexlab{}.
\newblock \showarticletitle{Detection and visualization of misleading content
  on Twitter}.
\newblock \bibinfo{journal}{\emph{International Journal of Multimedia
  Information Retrieval}} \bibinfo{volume}{7}, \bibinfo{number}{1}
  (\bibinfo{year}{2018}), \bibinfo{pages}{71--86}.
\newblock


\bibitem[Cao et~al\mbox{.}(2020)]%
        {Leveraging_5}
\bibfield{author}{\bibinfo{person}{Juan Cao}, \bibinfo{person}{Peng Qi},
  \bibinfo{person}{Qiang Sheng}, \bibinfo{person}{Tianyun Yang},
  \bibinfo{person}{Junbo Guo}, {and} \bibinfo{person}{Jintao Li}.}
  \bibinfo{year}{2020}\natexlab{}.
\newblock \showarticletitle{Exploring the role of visual content in fake news
  detection}.
\newblock \bibinfo{journal}{\emph{Disinformation, Misinformation, and Fake News
  in Social Media}} (\bibinfo{year}{2020}), \bibinfo{pages}{141--161}.
\newblock


\bibitem[Chen et~al\mbox{.}(2021)]%
        {MVSS}
\bibfield{author}{\bibinfo{person}{Xinru Chen}, \bibinfo{person}{Chengbo Dong},
  \bibinfo{person}{Jiaqi Ji}, \bibinfo{person}{Juan Cao}, {and}
  \bibinfo{person}{Xirong Li}.} \bibinfo{year}{2021}\natexlab{}.
\newblock \showarticletitle{Image Manipulation Detection by Multi-View
  Multi-Scale Supervision}. In \bibinfo{booktitle}{\emph{Proceedings of the
  IEEE/CVF International Conference on Computer Vision}}.
  \bibinfo{pages}{14185--14193}.
\newblock


\bibitem[Chen et~al\mbox{.}(2022)]%
        {WWW}
\bibfield{author}{\bibinfo{person}{Yixuan Chen}, \bibinfo{person}{Dongsheng
  Li}, \bibinfo{person}{Peng Zhang}, \bibinfo{person}{Jie Sui},
  \bibinfo{person}{Qin Lv}, \bibinfo{person}{Lu Tun}, {and} \bibinfo{person}{Li
  Shang}.} \bibinfo{year}{2022}\natexlab{}.
\newblock \showarticletitle{Cross-modal Ambiguity Learning for Multimodal Fake
  News Detection}. In \bibinfo{booktitle}{\emph{Proceedings of the ACM Web
  Conference 2022}}. \bibinfo{pages}{2897--2905}.
\newblock


\bibitem[Conde and Turgutlu(2021)]%
        {CLIP-Art}
\bibfield{author}{\bibinfo{person}{Marcos~V Conde} {and} \bibinfo{person}{Kerem
  Turgutlu}.} \bibinfo{year}{2021}\natexlab{}.
\newblock \showarticletitle{CLIP-Art: contrastive pre-training for fine-grained
  art classification}. In \bibinfo{booktitle}{\emph{Proceedings of the IEEE/CVF
  Conference on Computer Vision and Pattern Recognition}}.
  \bibinfo{pages}{3956--3960}.
\newblock


\bibitem[Conroy et~al\mbox{.}(2015)]%
        {Leveraging_6}
\bibfield{author}{\bibinfo{person}{Nadia~K Conroy}, \bibinfo{person}{Victoria~L
  Rubin}, {and} \bibinfo{person}{Yimin Chen}.} \bibinfo{year}{2015}\natexlab{}.
\newblock \showarticletitle{Automatic deception detection: Methods for finding
  fake news}.
\newblock \bibinfo{journal}{\emph{Proceedings of the association for
  information science and technology}} \bibinfo{volume}{52},
  \bibinfo{number}{1} (\bibinfo{year}{2015}), \bibinfo{pages}{1--4}.
\newblock


\bibitem[Dancette et~al\mbox{.}(2021)]%
        {VQA}
\bibfield{author}{\bibinfo{person}{Corentin Dancette}, \bibinfo{person}{Remi
  Cadene}, \bibinfo{person}{Damien Teney}, {and} \bibinfo{person}{Matthieu
  Cord}.} \bibinfo{year}{2021}\natexlab{}.
\newblock \showarticletitle{Beyond question-based biases: Assessing multimodal
  shortcut learning in visual question answering}. In
  \bibinfo{booktitle}{\emph{Proceedings of the IEEE/CVF International
  Conference on Computer Vision}}. \bibinfo{pages}{1574--1583}.
\newblock


\bibitem[Devlin et~al\mbox{.}(2018)]%
        {BERT}
\bibfield{author}{\bibinfo{person}{Jacob Devlin}, \bibinfo{person}{Ming-Wei
  Chang}, \bibinfo{person}{Kenton Lee}, {and} \bibinfo{person}{Kristina
  Toutanova}.} \bibinfo{year}{2018}\natexlab{}.
\newblock \showarticletitle{Bert: Pre-training of deep bidirectional
  transformers for language understanding}.
\newblock \bibinfo{journal}{\emph{arXiv preprint arXiv:1810.04805}}
  (\bibinfo{year}{2018}).
\newblock


\bibitem[Guo et~al\mbox{.}(2018)]%
        {guo2018rumor}
\bibfield{author}{\bibinfo{person}{Han Guo}, \bibinfo{person}{Juan Cao},
  \bibinfo{person}{Yazi Zhang}, \bibinfo{person}{Junbo Guo}, {and}
  \bibinfo{person}{Jintao Li}.} \bibinfo{year}{2018}\natexlab{}.
\newblock \showarticletitle{Rumor detection with hierarchical social attention
  network}. In \bibinfo{booktitle}{\emph{Proceedings of the 27th ACM
  international conference on information and knowledge management}}.
  \bibinfo{pages}{943--951}.
\newblock


\bibitem[Han et~al\mbox{.}(2021)]%
        {han2021fighting}
\bibfield{author}{\bibinfo{person}{Bing Han}, \bibinfo{person}{Xiaoguang Han},
  \bibinfo{person}{Hua Zhang}, \bibinfo{person}{Jingzhi Li}, {and}
  \bibinfo{person}{Xiaochun Cao}.} \bibinfo{year}{2021}\natexlab{}.
\newblock \showarticletitle{Fighting fake news: two stream network for deepfake
  detection via learnable SRM}.
\newblock \bibinfo{journal}{\emph{IEEE Transactions on Biometrics, Behavior,
  and Identity Science}} \bibinfo{volume}{3}, \bibinfo{number}{3}
  (\bibinfo{year}{2021}), \bibinfo{pages}{320--331}.
\newblock


\bibitem[He et~al\mbox{.}(2016a)]%
        {RESNET}
\bibfield{author}{\bibinfo{person}{Kaiming He}, \bibinfo{person}{Xiangyu
  Zhang}, \bibinfo{person}{Shaoqing Ren}, {and} \bibinfo{person}{Jian Sun}.}
  \bibinfo{year}{2016}\natexlab{a}.
\newblock \showarticletitle{Deep residual learning for image recognition}. In
  \bibinfo{booktitle}{\emph{Proceedings of the IEEE conference on computer
  vision and pattern recognition}}. \bibinfo{pages}{770--778}.
\newblock


\bibitem[He et~al\mbox{.}(2016b)]%
        {he2016deep}
\bibfield{author}{\bibinfo{person}{Kaiming He}, \bibinfo{person}{Xiangyu
  Zhang}, \bibinfo{person}{Shaoqing Ren}, {and} \bibinfo{person}{Jian Sun}.}
  \bibinfo{year}{2016}\natexlab{b}.
\newblock \showarticletitle{Deep residual learning for image recognition}. In
  \bibinfo{booktitle}{\emph{Proceedings of the IEEE conference on computer
  vision and pattern recognition}}. \bibinfo{pages}{770--778}.
\newblock


\bibitem[Hendrycks and Gimpel(2016)]%
        {GELU}
\bibfield{author}{\bibinfo{person}{Dan Hendrycks} {and} \bibinfo{person}{Kevin
  Gimpel}.} \bibinfo{year}{2016}\natexlab{}.
\newblock \showarticletitle{Gaussian error linear units (gelus)}.
\newblock \bibinfo{journal}{\emph{arXiv preprint arXiv:1606.08415}}
  (\bibinfo{year}{2016}).
\newblock


\bibitem[Hu et~al\mbox{.}(2018)]%
        {hu2018squeeze}
\bibfield{author}{\bibinfo{person}{Jie Hu}, \bibinfo{person}{Li Shen}, {and}
  \bibinfo{person}{Gang Sun}.} \bibinfo{year}{2018}\natexlab{}.
\newblock \showarticletitle{Squeeze-and-excitation networks}. In
  \bibinfo{booktitle}{\emph{Proceedings of the IEEE conference on computer
  vision and pattern recognition}}. \bibinfo{pages}{7132--7141}.
\newblock


\bibitem[Ioffe and Szegedy(2015)]%
        {BN}
\bibfield{author}{\bibinfo{person}{Sergey Ioffe} {and}
  \bibinfo{person}{Christian Szegedy}.} \bibinfo{year}{2015}\natexlab{}.
\newblock \showarticletitle{Batch normalization: Accelerating deep network
  training by reducing internal covariate shift}. In
  \bibinfo{booktitle}{\emph{International Conference on Machine Learning
  (ICML)}}. \bibinfo{pages}{448--456}.
\newblock


\bibitem[Jawahar et~al\mbox{.}(2019)]%
        {jawahar2019does}
\bibfield{author}{\bibinfo{person}{Ganesh Jawahar},
  \bibinfo{person}{Beno{\^\i}t Sagot}, {and} \bibinfo{person}{Djam{\'e}
  Seddah}.} \bibinfo{year}{2019}\natexlab{}.
\newblock \showarticletitle{What does BERT learn about the structure of
  language?}. In \bibinfo{booktitle}{\emph{ACL 2019-57th Annual Meeting of the
  Association for Computational Linguistics}}.
\newblock


\bibitem[Jin et~al\mbox{.}(2017)]%
        {jin2017multimodal}
\bibfield{author}{\bibinfo{person}{Zhiwei Jin}, \bibinfo{person}{Juan Cao},
  \bibinfo{person}{Han Guo}, \bibinfo{person}{Yongdong Zhang}, {and}
  \bibinfo{person}{Jiebo Luo}.} \bibinfo{year}{2017}\natexlab{}.
\newblock \showarticletitle{Multimodal fusion with recurrent neural networks
  for rumor detection on microblogs}. In \bibinfo{booktitle}{\emph{Proceedings
  of the 25th ACM international conference on Multimedia}}.
  \bibinfo{pages}{795--816}.
\newblock


\bibitem[Johnson(2012)]%
        {johnson2012google}
\bibfield{author}{\bibinfo{person}{Gregory Johnson}.}
  \bibinfo{year}{2012}\natexlab{}.
\newblock \showarticletitle{Google Translate http://translate. google. com}.
\newblock \bibinfo{journal}{\emph{Technical Services Quarterly}}
  \bibinfo{volume}{29}, \bibinfo{number}{2} (\bibinfo{year}{2012}),
  \bibinfo{pages}{165--165}.
\newblock


\bibitem[Khattar et~al\mbox{.}(2019)]%
        {MVAE}
\bibfield{author}{\bibinfo{person}{Dhruv Khattar},
  \bibinfo{person}{Jaipal~Singh Goud}, \bibinfo{person}{Manish Gupta}, {and}
  \bibinfo{person}{Vasudeva Varma}.} \bibinfo{year}{2019}\natexlab{}.
\newblock \showarticletitle{Mvae: Multimodal variational autoencoder for fake
  news detection}. In \bibinfo{booktitle}{\emph{The world wide web
  conference}}. \bibinfo{pages}{2915--2921}.
\newblock


\bibitem[Kingma and Ba(2014)]%
        {kingma2014adam}
\bibfield{author}{\bibinfo{person}{Diederik~P Kingma} {and}
  \bibinfo{person}{Jimmy Ba}.} \bibinfo{year}{2014}\natexlab{}.
\newblock \showarticletitle{Adam: A method for stochastic optimization}.
\newblock \bibinfo{journal}{\emph{arXiv preprint arXiv:1412.6980}}
  (\bibinfo{year}{2014}).
\newblock


\bibitem[Li et~al\mbox{.}(2021)]%
        {li2021entity}
\bibfield{author}{\bibinfo{person}{Peiguang Li}, \bibinfo{person}{Xian Sun},
  \bibinfo{person}{Hongfeng Yu}, \bibinfo{person}{Yu Tian},
  \bibinfo{person}{Fanglong Yao}, {and} \bibinfo{person}{Guangluan Xu}.}
  \bibinfo{year}{2021}\natexlab{}.
\newblock \showarticletitle{Entity-Oriented Multi-Modal Alignment and Fusion
  Network for Fake News Detection}.
\newblock \bibinfo{journal}{\emph{IEEE Transactions on Multimedia}}
  (\bibinfo{year}{2021}).
\newblock


\bibitem[Lin et~al\mbox{.}(2019)]%
        {lin2019detecting}
\bibfield{author}{\bibinfo{person}{Jun Lin}, \bibinfo{person}{Glenna
  Tremblay-Taylor}, \bibinfo{person}{Guanyi Mou}, \bibinfo{person}{Di You},
  {and} \bibinfo{person}{Kyumin Lee}.} \bibinfo{year}{2019}\natexlab{}.
\newblock \showarticletitle{Detecting fake news articles}. In
  \bibinfo{booktitle}{\emph{2019 IEEE International Conference on Big Data (Big
  Data)}}. IEEE, \bibinfo{pages}{3021--3025}.
\newblock


\bibitem[Lu et~al\mbox{.}(2019)]%
        {vilbert}
\bibfield{author}{\bibinfo{person}{Jiasen Lu}, \bibinfo{person}{Dhruv Batra},
  \bibinfo{person}{Devi Parikh}, {and} \bibinfo{person}{Stefan Lee}.}
  \bibinfo{year}{2019}\natexlab{}.
\newblock \showarticletitle{Vilbert: Pretraining task-agnostic visiolinguistic
  representations for vision-and-language tasks}.
\newblock \bibinfo{journal}{\emph{Advances in neural information processing
  systems}}  \bibinfo{volume}{32} (\bibinfo{year}{2019}).
\newblock


\bibitem[Nan et~al\mbox{.}(2021)]%
        {MDFEND}
\bibfield{author}{\bibinfo{person}{Qiong Nan}, \bibinfo{person}{Juan Cao},
  \bibinfo{person}{Yongchun Zhu}, \bibinfo{person}{Yanyan Wang}, {and}
  \bibinfo{person}{Jintao Li}.} \bibinfo{year}{2021}\natexlab{}.
\newblock \showarticletitle{MDFEND: Multi-domain Fake News Detection}. In
  \bibinfo{booktitle}{\emph{Proceedings of the 30th ACM International
  Conference on Information \& Knowledge Management}}.
  \bibinfo{pages}{3343--3347}.
\newblock


\bibitem[Nichol et~al\mbox{.}(2021)]%
        {Glide}
\bibfield{author}{\bibinfo{person}{Alex Nichol}, \bibinfo{person}{Prafulla
  Dhariwal}, \bibinfo{person}{Aditya Ramesh}, \bibinfo{person}{Pranav Shyam},
  \bibinfo{person}{Pamela Mishkin}, \bibinfo{person}{Bob McGrew},
  \bibinfo{person}{Ilya Sutskever}, {and} \bibinfo{person}{Mark Chen}.}
  \bibinfo{year}{2021}\natexlab{}.
\newblock \showarticletitle{Glide: Towards photorealistic image generation and
  editing with text-guided diffusion models}.
\newblock \bibinfo{journal}{\emph{arXiv preprint arXiv:2112.10741}}
  (\bibinfo{year}{2021}).
\newblock


\bibitem[Potthast et~al\mbox{.}(2017)]%
        {Leveraging_27}
\bibfield{author}{\bibinfo{person}{Martin Potthast}, \bibinfo{person}{Johannes
  Kiesel}, \bibinfo{person}{Kevin Reinartz}, \bibinfo{person}{Janek
  Bevendorff}, {and} \bibinfo{person}{Benno Stein}.}
  \bibinfo{year}{2017}\natexlab{}.
\newblock \showarticletitle{A stylometric inquiry into hyperpartisan and fake
  news}.
\newblock \bibinfo{journal}{\emph{arXiv preprint arXiv:1702.05638}}
  (\bibinfo{year}{2017}).
\newblock


\bibitem[Qi et~al\mbox{.}(2021)]%
        {Entity-Enhanced}
\bibfield{author}{\bibinfo{person}{Peng Qi}, \bibinfo{person}{Juan Cao},
  \bibinfo{person}{Xirong Li}, \bibinfo{person}{Huan Liu},
  \bibinfo{person}{Qiang Sheng}, \bibinfo{person}{Xiaoyue Mi},
  \bibinfo{person}{Qin He}, \bibinfo{person}{Yongbiao Lv},
  \bibinfo{person}{Chenyang Guo}, {and} \bibinfo{person}{Yingchao Yu}.}
  \bibinfo{year}{2021}\natexlab{}.
\newblock \showarticletitle{Improving Fake News Detection by Using an
  Entity-enhanced Framework to Fuse Diverse Multimodal Clues}. In
  \bibinfo{booktitle}{\emph{Proceedings of the 29th ACM International
  Conference on Multimedia}}. \bibinfo{pages}{1212--1220}.
\newblock


\bibitem[Qi et~al\mbox{.}(2019)]%
        {ICASSP-1}
\bibfield{author}{\bibinfo{person}{Peng Qi}, \bibinfo{person}{Juan Cao},
  \bibinfo{person}{Tianyun Yang}, \bibinfo{person}{Junbo Guo}, {and}
  \bibinfo{person}{Jintao Li}.} \bibinfo{year}{2019}\natexlab{}.
\newblock \showarticletitle{Exploiting multi-domain visual information for fake
  news detection}. In \bibinfo{booktitle}{\emph{2019 IEEE International
  Conference on Data Mining (ICDM)}}. IEEE, \bibinfo{pages}{518--527}.
\newblock


\bibitem[Radford et~al\mbox{.}(2021)]%
        {CLIP}
\bibfield{author}{\bibinfo{person}{Alec Radford}, \bibinfo{person}{Jong~Wook
  Kim}, \bibinfo{person}{Chris Hallacy}, \bibinfo{person}{Aditya Ramesh},
  \bibinfo{person}{Gabriel Goh}, \bibinfo{person}{Sandhini Agarwal},
  \bibinfo{person}{Girish Sastry}, \bibinfo{person}{Amanda Askell},
  \bibinfo{person}{Pamela Mishkin}, \bibinfo{person}{Jack Clark},
  {et~al\mbox{.}}} \bibinfo{year}{2021}\natexlab{}.
\newblock \showarticletitle{Learning transferable visual models from natural
  language supervision}. In \bibinfo{booktitle}{\emph{International Conference
  on Machine Learning}}. PMLR, \bibinfo{pages}{8748--8763}.
\newblock


\bibitem[Raffel et~al\mbox{.}(2019)]%
        {raffel2019exploring}
\bibfield{author}{\bibinfo{person}{Colin Raffel}, \bibinfo{person}{Noam
  Shazeer}, \bibinfo{person}{Adam Roberts}, \bibinfo{person}{Katherine Lee},
  \bibinfo{person}{Sharan Narang}, \bibinfo{person}{Michael Matena},
  \bibinfo{person}{Yanqi Zhou}, \bibinfo{person}{Wei Li}, {and}
  \bibinfo{person}{Peter~J Liu}.} \bibinfo{year}{2019}\natexlab{}.
\newblock \showarticletitle{Exploring the limits of transfer learning with a
  unified text-to-text transformer}.
\newblock \bibinfo{journal}{\emph{arXiv preprint arXiv:1910.10683}}
  (\bibinfo{year}{2019}).
\newblock


\bibitem[Shu et~al\mbox{.}(2020a)]%
        {shu2020fakenewsnet}
\bibfield{author}{\bibinfo{person}{Kai Shu}, \bibinfo{person}{Deepak
  Mahudeswaran}, \bibinfo{person}{Suhang Wang}, \bibinfo{person}{Dongwon Lee},
  {and} \bibinfo{person}{Huan Liu}.} \bibinfo{year}{2020}\natexlab{a}.
\newblock \showarticletitle{Fakenewsnet: A data repository with news content,
  social context, and spatiotemporal information for studying fake news on
  social media}.
\newblock \bibinfo{journal}{\emph{Big data}} \bibinfo{volume}{8},
  \bibinfo{number}{3} (\bibinfo{year}{2020}), \bibinfo{pages}{171--188}.
\newblock


\bibitem[Shu et~al\mbox{.}(2020b)]%
        {shu2020leveraging}
\bibfield{author}{\bibinfo{person}{Kai Shu}, \bibinfo{person}{Guoqing Zheng},
  \bibinfo{person}{Yichuan Li}, \bibinfo{person}{Subhabrata Mukherjee},
  \bibinfo{person}{Ahmed~Hassan Awadallah}, \bibinfo{person}{Scott Ruston},
  {and} \bibinfo{person}{Huan Liu}.} \bibinfo{year}{2020}\natexlab{b}.
\newblock \showarticletitle{Leveraging multi-source weak social supervision for
  early detection of fake news}.
\newblock \bibinfo{journal}{\emph{arXiv preprint arXiv:2004.01732}}
  (\bibinfo{year}{2020}).
\newblock


\bibitem[Singhal et~al\mbox{.}(2020)]%
        {SpotFake}
\bibfield{author}{\bibinfo{person}{Shivangi Singhal}, \bibinfo{person}{Anubha
  Kabra}, \bibinfo{person}{Mohit Sharma}, \bibinfo{person}{Rajiv~Ratn Shah},
  \bibinfo{person}{Tanmoy Chakraborty}, {and} \bibinfo{person}{Ponnurangam
  Kumaraguru}.} \bibinfo{year}{2020}\natexlab{}.
\newblock \showarticletitle{Spotfake+: A multimodal framework for fake news
  detection via transfer learning (student abstract)}. In
  \bibinfo{booktitle}{\emph{Proceedings of the AAAI Conference on Artificial
  Intelligence}}, Vol.~\bibinfo{volume}{34}. \bibinfo{pages}{13915--13916}.
\newblock


\bibitem[Singhal et~al\mbox{.}(2022)]%
        {singhal2022leveraging}
\bibfield{author}{\bibinfo{person}{Shivangi Singhal}, \bibinfo{person}{Tanisha
  Pandey}, \bibinfo{person}{Saksham Mrig}, \bibinfo{person}{Rajiv~Ratn Shah},
  {and} \bibinfo{person}{Ponnurangam Kumaraguru}.}
  \bibinfo{year}{2022}\natexlab{}.
\newblock \showarticletitle{Leveraging Intra and Inter Modality Relationship
  for Multimodal Fake News Detection}.
\newblock  (\bibinfo{year}{2022}).
\newblock


\bibitem[Singhal et~al\mbox{.}(2019)]%
        {singhal2019spotfake}
\bibfield{author}{\bibinfo{person}{Shivangi Singhal},
  \bibinfo{person}{Rajiv~Ratn Shah}, \bibinfo{person}{Tanmoy Chakraborty},
  \bibinfo{person}{Ponnurangam Kumaraguru}, {and} \bibinfo{person}{Shin'ichi
  Satoh}.} \bibinfo{year}{2019}\natexlab{}.
\newblock \showarticletitle{Spotfake: A multi-modal framework for fake news
  detection}. In \bibinfo{booktitle}{\emph{2019 IEEE fifth international
  conference on multimedia big data (BigMM)}}. IEEE, \bibinfo{pages}{39--47}.
\newblock


\bibitem[Van~der Maaten and Hinton(2008)]%
        {van2008visualizing}
\bibfield{author}{\bibinfo{person}{Laurens Van~der Maaten} {and}
  \bibinfo{person}{Geoffrey Hinton}.} \bibinfo{year}{2008}\natexlab{}.
\newblock \showarticletitle{Visualizing data using t-SNE.}
\newblock \bibinfo{journal}{\emph{Journal of machine learning research}}
  \bibinfo{volume}{9}, \bibinfo{number}{11} (\bibinfo{year}{2008}).
\newblock


\bibitem[Wang et~al\mbox{.}(2018)]%
        {EANN}
\bibfield{author}{\bibinfo{person}{Yaqing Wang}, \bibinfo{person}{Fenglong Ma},
  \bibinfo{person}{Zhiwei Jin}, \bibinfo{person}{Ye Yuan},
  \bibinfo{person}{Guangxu Xun}, \bibinfo{person}{Kishlay Jha},
  \bibinfo{person}{Lu Su}, {and} \bibinfo{person}{Jing Gao}.}
  \bibinfo{year}{2018}\natexlab{}.
\newblock \showarticletitle{Eann: Event adversarial neural networks for
  multi-modal fake news detection}. In \bibinfo{booktitle}{\emph{Proceedings of
  the 24th acm sigkdd international conference on knowledge discovery \& data
  mining}}. \bibinfo{pages}{849--857}.
\newblock


\bibitem[Wei et~al\mbox{.}(2021)]%
        {Hair-CLIP}
\bibfield{author}{\bibinfo{person}{Tianyi Wei}, \bibinfo{person}{Dongdong
  Chen}, \bibinfo{person}{Wenbo Zhou}, \bibinfo{person}{Jing Liao},
  \bibinfo{person}{Zhentao Tan}, \bibinfo{person}{Lu Yuan},
  \bibinfo{person}{Weiming Zhang}, {and} \bibinfo{person}{Nenghai Yu}.}
  \bibinfo{year}{2021}\natexlab{}.
\newblock \showarticletitle{Hairclip: Design your hair by text and reference
  image}.
\newblock \bibinfo{journal}{\emph{arXiv preprint arXiv:2112.05142}}
  (\bibinfo{year}{2021}).
\newblock


\bibitem[Wu et~al\mbox{.}(2021)]%
        {wu2021multimodal}
\bibfield{author}{\bibinfo{person}{Yang Wu}, \bibinfo{person}{Pengwei Zhan},
  \bibinfo{person}{Yunjian Zhang}, \bibinfo{person}{Liming Wang}, {and}
  \bibinfo{person}{Zhen Xu}.} \bibinfo{year}{2021}\natexlab{}.
\newblock \showarticletitle{Multimodal Fusion with Co-Attention Networks for
  Fake News Detection}. In \bibinfo{booktitle}{\emph{Findings of the
  Association for Computational Linguistics: ACL-IJCNLP 2021}}.
  \bibinfo{pages}{2560--2569}.
\newblock


\bibitem[Xue et~al\mbox{.}(2021)]%
        {xue2021detecting}
\bibfield{author}{\bibinfo{person}{Junxiao Xue}, \bibinfo{person}{Yabo Wang},
  \bibinfo{person}{Yichen Tian}, \bibinfo{person}{Yafei Li},
  \bibinfo{person}{Lei Shi}, {and} \bibinfo{person}{Lin Wei}.}
  \bibinfo{year}{2021}\natexlab{}.
\newblock \showarticletitle{Detecting fake news by exploring the consistency of
  multimodal data}.
\newblock \bibinfo{journal}{\emph{Information Processing \& Management}}
  \bibinfo{volume}{58}, \bibinfo{number}{5} (\bibinfo{year}{2021}),
  \bibinfo{pages}{102610}.
\newblock


\bibitem[Ye and Kovashka(2021)]%
        {VCR}
\bibfield{author}{\bibinfo{person}{Keren Ye} {and} \bibinfo{person}{Adriana
  Kovashka}.} \bibinfo{year}{2021}\natexlab{}.
\newblock \showarticletitle{A case study of the shortcut effects in visual
  commonsense reasoning}. In \bibinfo{booktitle}{\emph{Proceedings of the AAAI
  conference on artificial intelligence}}, Vol.~\bibinfo{volume}{35}.
  \bibinfo{pages}{3181--3189}.
\newblock


\bibitem[Zhang et~al\mbox{.}(2021)]%
        {Dual-Emotion}
\bibfield{author}{\bibinfo{person}{Xueyao Zhang}, \bibinfo{person}{Juan Cao},
  \bibinfo{person}{Xirong Li}, \bibinfo{person}{Qiang Sheng},
  \bibinfo{person}{Lei Zhong}, {and} \bibinfo{person}{Kai Shu}.}
  \bibinfo{year}{2021}\natexlab{}.
\newblock \showarticletitle{Mining dual emotion for fake news detection}. In
  \bibinfo{booktitle}{\emph{Proceedings of the Web Conference 2021}}.
  \bibinfo{pages}{3465--3476}.
\newblock


\bibitem[Zhou et~al\mbox{.}(2020)]%
        {zhou2020mathsf}
\bibfield{author}{\bibinfo{person}{Xinyi Zhou}, \bibinfo{person}{Jindi Wu},
  {and} \bibinfo{person}{Reza Zafarani}.} \bibinfo{year}{2020}\natexlab{}.
\newblock \showarticletitle{SAFE: Similarity-Aware Multi-modal Fake News
  Detection}. In \bibinfo{booktitle}{\emph{Pacific-Asia Conference on Knowledge
  Discovery and Data Mining}}. Springer, \bibinfo{pages}{354--367}.
\newblock


\bibitem[Zubiaga et~al\mbox{.}(2018)]%
        {FND-Survey}
\bibfield{author}{\bibinfo{person}{Arkaitz Zubiaga}, \bibinfo{person}{Ahmet
  Aker}, \bibinfo{person}{Kalina Bontcheva}, \bibinfo{person}{Maria Liakata},
  {and} \bibinfo{person}{Rob Procter}.} \bibinfo{year}{2018}\natexlab{}.
\newblock \showarticletitle{Detection and resolution of rumours in social
  media: A survey}.
\newblock \bibinfo{journal}{\emph{ACM Computing Surveys (CSUR)}}
  \bibinfo{volume}{51}, \bibinfo{number}{2} (\bibinfo{year}{2018}),
  \bibinfo{pages}{1--36}.
\newblock


\end{thebibliography}


\end{document}